\documentclass[conference]{IEEEtran}
\IEEEoverridecommandlockouts
\usepackage{xpatch}
\makeatletter
\patchcmd\@makecaption{\\}{.~}{}{\fail}
\makeatletter
\usepackage{cite}
\usepackage{amsmath,amssymb,amsfonts}
\usepackage{algorithmic}
\usepackage{graphicx}
\usepackage{textcomp}
\usepackage{xcolor}
\usepackage{multirow}
\usepackage{hyperref}
\usepackage{subfig}
\def\BibTeX{{\rm B\kern-.05em{\sc i\kern-.025em b}\kern-.08em
    T\kern-.1667em\lower.7ex\hbox{E}\kern-.125emX}}
\begin{document}

\title{Moving Forward: A Review of Autonomous Driving Software and Hardware Systems
}


\author{
\IEEEauthorblockN{
Xu Wang\IEEEauthorrefmark{1},
Mohammad Ali Maleki\IEEEauthorrefmark{1},
Muhammad Waqar Azhar\IEEEauthorrefmark{2},
Pedro Trancoso\IEEEauthorrefmark{1}
\IEEEauthorblockA{
\IEEEauthorrefmark{1} Chalmers University of Technology and University of Gothenburg,
\IEEEauthorrefmark{2} ZeroPoint Technologies\\
\IEEEauthorrefmark{1} \{xuwang, mohammad.ali.maleki, ppedro\}@chalmers.se}
\IEEEauthorrefmark{2} \{waqar.azhar\}@zptcorp.com}
}

\maketitle

\begin{abstract}
With their potential to significantly reduce traffic accidents, enhance road safety, optimize traffic flow, and decrease congestion, autonomous driving systems are a major focus of research and development in recent years. Beyond these immediate benefits, they offer long-term advantages in promoting sustainable transportation by reducing emissions and fuel consumption.
Achieving a high level of autonomy across diverse conditions requires a comprehensive understanding of the environment. This is accomplished by processing data from sensors such as cameras, radars, and LiDARs through a software stack that relies heavily on machine learning algorithms. These ML models demand significant computational resources and involve large-scale data movement, presenting challenges for hardware to execute them efficiently and at high speed.

In this survey, we first outline and highlight the key components of self-driving systems, covering input sensors, commonly used datasets, simulation platforms, and the software architecture. We then explore the underlying hardware platforms that support the execution of these software systems.

By presenting a comprehensive view of autonomous driving systems and their increasing demands, particularly for higher levels of autonomy, we analyze the performance and efficiency of scaled-up off-the-shelf GPU/CPU-based systems, emphasizing the challenges within the computational components. Through examples showcasing the diverse computational and memory requirements in the software stack, we demonstrate how more specialized hardware and processing closer to memory can enable more efficient execution with lower latency. Finally, based on current trends and future demands, we conclude by speculating what a future hardware platform for autonomous driving might look like.
\end{abstract}


\section{Introduction}
\label{sec:introduction}

Autonomous driving systems have gained significant attention because of their potential to: (1) enhance road safety by reducing traffic accidents caused by human error; (2) improve traffic efficiency and productivity by optimizing traffic flow; and (3) minimize environmental impact by optimizing driving patterns, which in turn lowers fuel consumption and promotes sustainability ~\cite{li2023evaluation}. 

The Society of Automotive Engineers (SAE) has developed a taxonomy for the different levels of autonomous driving ~\cite{autonomous-driving-levels}. There are six levels in this taxonomy, ranging from {\em No Automation} (level 0) up to {\em Full Automation} (level 5). Currently, available vehicles are certified as 
{\em Partial} or {\em Conditional Automation} (level 2 or 3, respectively). However, recent announcements and developments are positioning experimental and near-future products in the High Driving Automation category (level 4).

Towards this goal, recent advances in computer vision and deep learning (DL) models, as well as more efficient hardware, have enabled the evolution of autonomous driving systems. These innovations have improved the accuracy and effectiveness of decision-making in complex scenarios, such as urban environments, offering improvements over traditional rule-based systems~\cite{sana2023autonomous}.

\begin{figure}[htbp]
    \centering
    \includegraphics[width=0.45\textwidth]{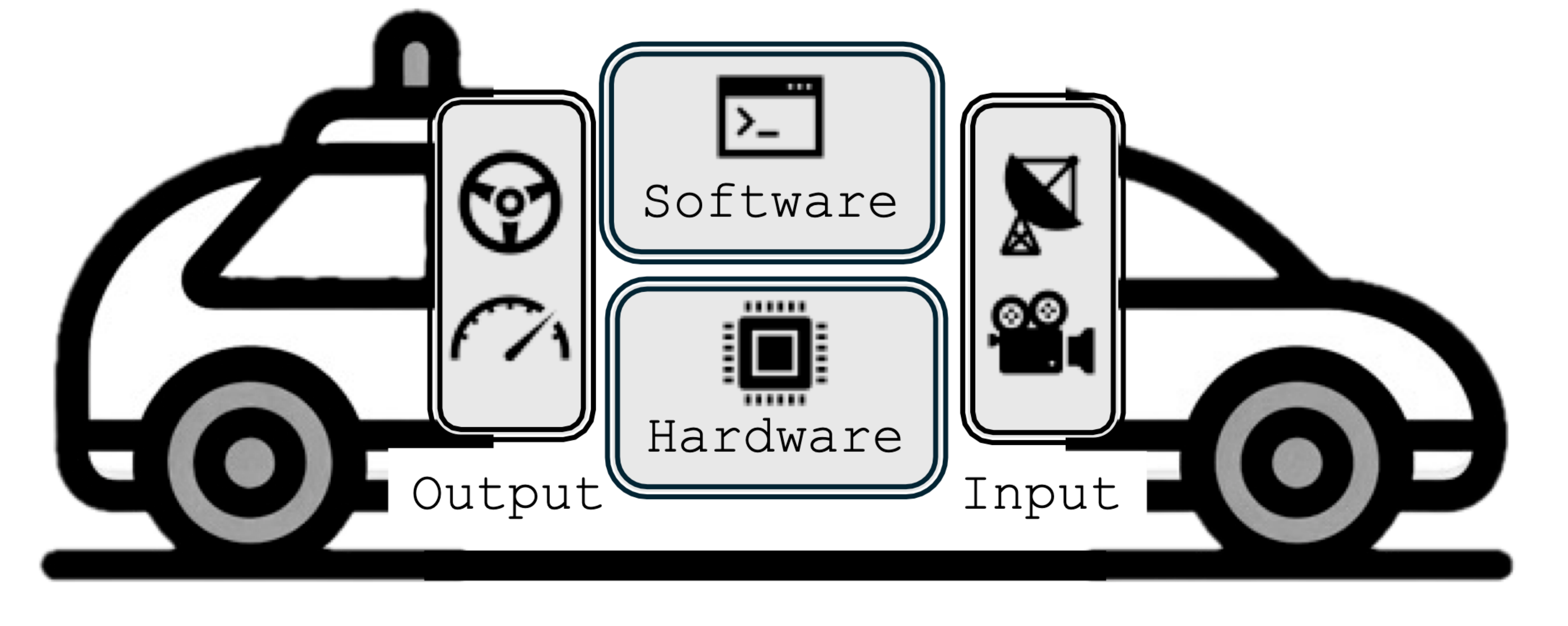}
    \caption{Autonomous driving system modules.}
    \label{fig:self-driving components}
\end{figure}

A high-level simplified view of an autonomous driving system is depicted in Figure~\ref{fig:self-driving components}. This system can be represented as a collection of four basic modules: {\em Input}, {\em Output}, {\em Software}, and {\em Hardware}. The {\em Input} module consists of different types of sensors (e.g. cameras) that are used to reconstruct the surrounding environment of the vehicle, as well as to provide its current location. The {\em Output} module includes actuators, such as steering, accelerator, and brakes, that given a certain decision, are able to influence the vehicle's movement.

The decisions are determined by the {\em Software} module. The core of this module is a complex DL model or set of DL models which take as input the information from the sensors, process it according to the trained parameters and produce decisions that are then sent to the actuators. Different autonomous driving DL models have been proposed. Yurtsever {\em et al.}~\cite{Yurtsever2019ASO} showed that DL autonomous driving systems typically can be divided into two categories: (1) generic modular systems and (2) end-to-end systems. 
initially, generic modular systems aimed to decompose the complex autonomous driving problem into smaller easier to tackle sub-problems, using a different specialized DL model for each of those problems. In contrast, end-to-end systems were proposed as a way to more closely model the way that humans approach the driving problem as a whole with a single complex model. This approach is preferred by most current solutions.

The last module is the {\em Hardware} which is composed of computer systems (e.g. CPU- or GPU-based system) capable of providing the necessary computational power to execute the above-mentioned DL model within the tight requirements for this application.

The combination of all these modules forms a complete solution for autonomous driving. Several such solutions have been proposed in the past. 
Boss~\cite{Urmson2008AutonomousDI} is the solution that won the 2007 DARPA Urban Challenge. IARA~\cite{Badue2019SelfDrivingCA} is the first Brazilian self-driving car solution.
Waymo~\cite{Waymo} is a self-driving car solution developed by Google.
Other relevant solutions are Apollo~\cite{Apollo} from Baidu and the solution offered by Tesla~\cite{TeslaAutopilot2023}. As mentioned before, the ultimate goal for autonomous driving is to progress from the existing solutions, which are at levels 2 to 4, towards achieving novel solutions for {\em Full Automation} at level 5. 

Several challenges need to be addressed in order to reach this goal. To achieve level 5, there is a need for better environmental information, requiring more input sensors of different types and higher accuracy~\cite{hurair2024environmental}. This immediately results in more complex and larger DL models capable of handling all this input information. In addition, these models will also become more complex as they are required to handle more corner cases. All of this will require a more powerful and at the same time more energy-efficient hardware platform to satisfy all the needs. 
The goal of this work is to present the challenges and directions that designers should consider for the development of the hardware accelerators needed to support level 5 autonomous driving. To set he stage for this, we present an extensive review of the state-of-the-art in terms of the most relevant DL models and hardware systems for autonomous driving, as well as emerging technologies for hardware accelerators. 

\section{Input Sensors And Evaluation Frameworks and Datasets}
\label{se:input/output}

Before delving into the details of autonomous vehicle's different modules, it's essential to provide some contextual information. The material presented in this review is based on comprehensive solutions for autonomous driving. Notable examples include Boss, which won the 2007 DARPA Urban Challenge ~\cite{Urmson2008AutonomousDI}; IARA ~\cite{Badue2019SelfDrivingCA}, the first Brazilian self-driving car; Waymo ~\cite{Waymo}, developed by Google; Apollo from Baidu ~\cite{Apollo}; and the solution offered by Tesla ~\cite{TeslaAutopilot2023}. These examples will serve as references when discussing these various modules.
Considering this, we provide a brief overview of the most relevant input sensors used in autonomous driving in Section~\ref{sec:input_sensor}. In Section ~\ref{sec:simulators_datasets}, we present the different available frameworks and datasets for testing and evaluation of autonomous driving systems.


\subsection{Input Sensors}
\label{sec:input_sensor}

Autonomous driving task is highly dependent on the understanding of the environment surrounding the vehicle. Different types of sensors are used to reconstruct the environment. Intuitively, a greater number and variety of sensors should result in a better perception of the environment, thereby enabling improved decision-making. At the same time, processing the vast amount of information provided by these sensors requires significant computational power. 
This creates an important trade-off that must be carefully examined, as it is currently considered a limiting factor in achieving high levels of driving automation.

In most cases, different types of sensors complement each other to provide a more comprehensive view of the environment. For instance, cameras, which are passive sensors, capture 2D image data and can be used in clusters to reconstruct a 3D view from 2D images. However, cameras are sensitive to changes in illumination and weather conditions, potentially producing poor-quality images in dark or rainy environments. Unlike passive cameras, LiDARs are less sensitive to illumination and weather conditions as they are active sensors that can emit and receive reflective infrared light. In addition, LiDARs can provide 3D data with depth information.\\
Camera and LiDAR fusion is gaining increasing interest because it leverages both the depth information from LiDAR point clouds and the color information from 2D images \cite{TSF-FCN}. Radars are another type of active sensor used in autonomous vehicles. Unlike LiDARs, radars emit radio waves rather than infrared light waves. While they can detect objects at greater distances, their accuracy is generally lower compared to LiDARs. GPS sensors are another type widely used in autonomous vehicles. GPS provides global location information, which is crucial for vehicle navigation. To improve localization accuracy, GPS data is often complemented with local information from an Inertial Measurement Unit (IMU). 

 Different self-driving systems employ various combinations of sensors. For instance, Boss 
 utilizes eleven LiDARs, five radars, and two cameras. IARA operates autonomously with a combination of LiDARs and cameras. Waymo's system employs multiple radars, LiDARs, and cameras. Apollo integrates LiDARs into its setup, while Tesla has opted for a camera-only self-driving system, eliminating the use of radar and ultrasonic sensors. Table~\ref{tab:sensors settings} summarizes the sensor configuration used by different systems.

\begin{table}[ht]
    \centering
        \caption{Sensor configuration of different self-driving systems}
    \begin{tabular}{c|c|c|c|c|c}
    \hline
    \textbf{System}  &  \textbf{Camera} & \textbf{LiDAR} & \textbf{Radar} & \textbf{GPS} & \textbf{IMU} \\ \hline
     Boss \cite{Urmson2008AutonomousDI}    &  2  & 11 & 5 & \checkmark & \checkmark \\ 
      IARA \cite{Badue2019SelfDrivingCA}     & 3 & 2 & - & \checkmark & \checkmark    \\ 
     Waymo \cite{Waymo}     & 9   & 4 & 6 & \checkmark & \checkmark    \\ 
     Apollo \cite{Apollo} & 2    & 4 & 2   & \checkmark & \checkmark       \\ 
     Tesla \cite{TeslaAutopilot2023} & 8    & - & -  & \checkmark & \checkmark    \\  \hline
     \end{tabular}
    \vspace{0.2cm}

    \label{tab:sensors settings}
\end{table}

\begin{figure*}[htbp]
\centering
   \begin{minipage}{0.9\textwidth}
   
   \begin{minipage}{0.45\textwidth}
   \centering
   \includegraphics[width=\textwidth]{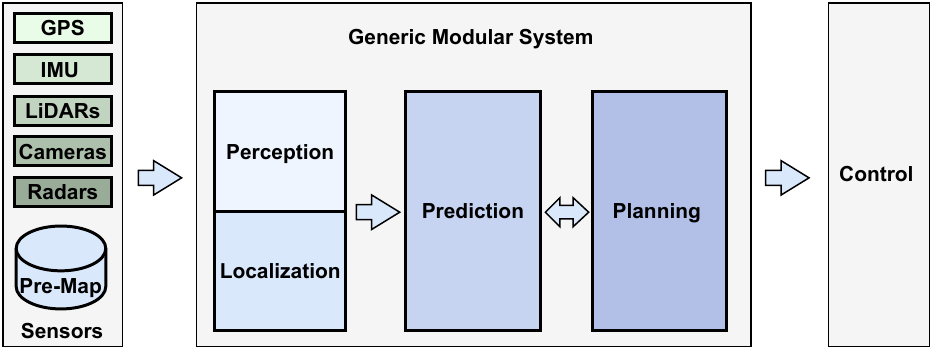}
   {\small (a)}
   \end{minipage}
   \hspace{0.05\textwidth}
   \begin{minipage}{0.45\textwidth}
   \centering
   \includegraphics[width=\textwidth]{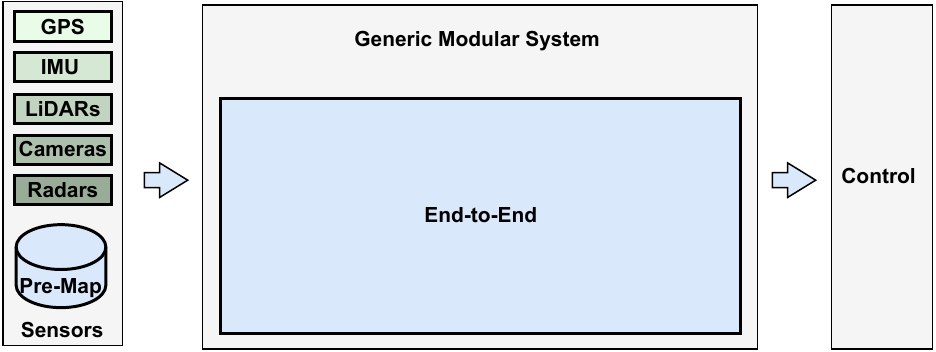}
   {\small (b)}
   \end{minipage}
       
   \end{minipage}

\caption{System diagram of: (a) generic modular system, (b) end-to-end system.}
\label{fig:system_diagram}
\end{figure*}

\subsection{Evaluation Frameworks and Datasets}
\label{sec:simulators_datasets}

Before deploying self-driving systems, it is crucial that these systems undergo rigorous training and testing processes. Training involves providing deep learning (DL) models with labeled data from comprehensive datasets, allowing them to learn and adapt to various driving conditions and scenarios. Once the models are trained, validation and testing should be conducted to assess their performance and reliability, utilizing simulators and datasets. 

In the generic modular systems, the complex self-driving problem is divided into distinct tasks (see Section~\ref{sec:software_systems}) each of which can be trained independently with inputs tailored to its specific function.The most common types of training inputs for these modules are images and point clouds, which capture the environment in 2D and 3D forms respectively. Regarding training image-based models, there are various datasets available, thanks to the advancements in computer vision. \textit{PASCAL Visual Object Classes (VOC)} project \cite{Everingham2010ThePV}, particularly the VOC2007 and VOC2012, are well-known datasets and challenges for object detection, semantic segmentation, and image classification. They contain 20 predefined class images, including vehicles, bicycles, animals, etc, and each image has been annotated with information such as the belonging objects for classification, bound box coordinates for object detection, and pixel-level details for segmentation.

Datasets such as \textit{ImageNet} \cite{Deng2009ImageNetAL}, \textit{COCO} \cite{Lin2014MicrosoftCC}, and \textit{CIFAR-10} and \textit{CIFAR-100} \cite{Krizhevsky2009LearningML} are also commonly used in computer vision and can be utilized to train the vision modules in self-driving systems. The \textit{Cityscapes} \cite{Cordts2016TheCD} dataset which includes video sequences of street scenes from 50 different cities with both fine and coarse annotations, is particularly useful for training the vision system of automated driving systems, especially for segmentation tasks.

As LiDAR becomes increasingly integrated into self-driving systems, a variety of datasets have been developed to provide the necessary 3D point cloud data for training and validating these systems. \textit{Karlsruhe Institute of Technology and Toyota Technological Institute (KITTI)} \cite{Geisberger2012ExactRI} is a popular dataset for self-driving, especially focusing on 3D object detection and 3D tracking. It contains both the LiDAR point clouds and the camera images, enabling the training and evaluation of LiDAR-based algorithms, such as VoxelNet \cite{VoxelNet} and PointPillar \cite{Lang2018PointPillarsFE}, as well as LiDAR-Camera fusion algorithms like MV3D \cite{MV3D} and AVOD \cite{AVOD}.

The \textit{Waymo open dataset} includes both a perception dataset \cite{Sun2019ScalabilityIP} and a motion dataset \cite{Ettinger2021LargeSI, Chen2023WOMDLiDARRS}. The perception dataset contains the high-resolution camera and LiDAR data with detailed labels like bounding boxes and segmentation, ideal for training and validating perception models. The motion dataset focuses on dynamic objects, providing tracking IDs, labels, and bounding boxes, along with map information from six cities, making it valuable for developing self-driving systems in diverse urban environments.\\ 

Integrated datasets and simulators have been developed to facilitate the deployment of self-driving systems. \textit{Apollo} \cite{Apollo} and \textit{Autoware} \cite{autowarefoundation} are generic modular platforms that offer full-stack development framework. Specifically,  \textit{Apollo} \cite{Apollo}, developed by Baidu, provides the \textit{ApolloScape} \cite{Wang2018TheAO} dataset, which offers more detailed label information compared to KITTI \cite{Geisberger2012ExactRI} and Cityscapes \cite{Cordts2016TheCD}. Additionally,  \textit{Apollo} includes a visualization and simulation tool, \textit{Dreamview}, which enhances user interaction with various modules, allowing for the deployment of pre-trained models.

\textit{Autoware} \cite{autowarefoundation} comprises of various software modules for self-driving. It also can be integrated with an LGSVL simulator \cite{rong2020lgsvl} to validate the model deployment. \textit{Car Learning to Act (CARLA)} \cite{Dosovitskiy17} is another open-source simulator for autonomous driving in urban environments. It can be deployed to train and evaluate the models for self-driving, including perception and control algorithms. CARLA also provides diverse kinds of city layouts and different kinds of vehicles and buildings. The simulator provides different environmental conditions, weather and time, and different sensor suites. According to the setup, the models can be trained and validated within this simulator. Three self-driving approaches are evaluated in CARLA, including generic modular pipeline, end-to-end imitation learning, and end-to-end reinforcement learning.

\section{Software Systems}
\label{sec:software_systems}
With the sensors gathering data from the surrounding environment, the challenge now lies in interpreting this information and, based on that understanding, making the appropriate control decisions and reactions. To tackle this issue, various autonomous driving software solutions have been proposed. 
Yurtsever {\em et al.}~\cite{Yurtsever2019ASO} showed that 
DL autonomous driving systems typically can be divided into two categories: (1) generic modular systems and (2) end-to-end systems.
These two categories are depicted in Figure~\ref{fig:system_diagram} and will be discussed in the following subsections.

\subsection{Generic Modular Software Systems}
\label{sec:generic_modular_systems}

The generic modular system breaks down the self-driving process into multiple modules with each module dedicated to a specific function. The granularity with which these tasks are divided can vary among different self-driving systems, but typically they are segmented into five stages: perception, localization,  prediction, planning, and control, see Figure~\ref{fig:system_diagram}(a).

The perception module detects and interprets its surrounding environment by utilizing the inputs from different sensors, such as LiDAR and cameras. This information is then used to identify traffic lights and obstacles, such as pedestrians and other vehicles. Localization is achieved by interpreting data from GPS and IMU to generate the location information of the ego vehicle. Armed with an understanding of itself and its surroundings, the prediction module then forecasts the trajectories of nearby obstacles. Subsequently, the planning module determines the optimal path and actions based on the perceived and predicted information, while ensuring compliance with traffic rules.\bigskip

\subsubsection{Perception}
\hfill \break
The Perception module processes raw sensor inputs, including data from cameras and LiDARs/Radars, to generate surrounding information, such as the presence of obstacles and the status of traffic lights. Then, this information is relayed to the downstream modules, such as prediction and planning, to facilitate informed decision-making and control.

\textbf{Object detection:} object detection involves determining the location and dimensions of target objects within a given context. This task can be categorized into two primary classes: 2D object detection and 3D object detection. Moreover, in these two main categories, there are approaches that make use of the combination of camera and LiDAR to enhance the accuracy of object detection.

In 2D object detection, key models include Faster R-CNN \cite{Ren2015FasterRCNN}, which utilizes a region proposal network for the Fast R-CNN detector. YOLO \cite{YOLO}, a one-stage network, predicts bounding boxes and class probabilities without a separate classifier, resulting in shorter inference times compared to Faster R-CNN. Similarly, SSD \cite{Liu2015SSDSS} is another one-stage network that offers faster speed and higher accuracy than YOLO.

Given the sensitivity of cameras to variations in illumination and weather conditions, 3D object detection methods using LiDAR have been explored. Notable models include VoxelNet \cite{VoxelNet}, which operates directly on sparse 3D points without hand-crafted feature representations, and SECOND \cite{SECOND}, which uses sparse convolution networks to achieve short inference times. PointNet \cite{PointNet} and PointNet++ \cite{PointNet++} process point clouds directly from LiDAR without voxelization, which enables diverse 3D object detection algorithms such as PointPillar \cite{Lang2018PointPillarsFE}.

Enhancing object detection accuracy can be achieved by integrating the detailed visual information from 2D images with the complementary depth information provided by 3D point clouds. F-PointNet \cite{FrustumPF} builds upon an image-based CNN and PointNet variants to directly locate 3D objects on point clouds. Compared with the F-PointNet \cite{FrustumPF}, which fuses two modalities at an early stage, MV3D \cite{MV3D} employs a later-stage fusing strategy. MV3D employs three primary inputs: a bird view and a front view from LiDAR, along with 2D image from the camera. Similar to the \cite{MV3D}, the AVOD \cite{AVOD} also adopts a later-fusion strategy. It utilizes two VGG \cite{Simonyan2014VeryDC} networks to extract the 2D and 3D features, respectively. 
Transformer models, as described in \cite{Vaswani2017AttentionIA}, have the potential to integrate various modalities and provide a holistic view.
Transformers such as TransFusion \cite{Bai2022TransFusionRL} and UVTR \cite{Li2022UnifyingVR} offer a potential for integrating various modalities \cite{Vaswani2017AttentionIA} and managing long sequences. 
A more comprehensive and in-depth discussion about object detection can be found in \cite{Mao20223DOD}. Table ~\ref{tab:object_detection} summarizes the performance of different object detection methods discussed. \\

\begin{table}[ht]
    \centering
        \vspace{-0.2 cm}
    \caption{Performance of different object detection methods  with their sensor combination. mAP refers to mean Average Precision}
        \vspace{-0.2 cm}
    \begin{tabular}{c|c|c|c|c}
    \hline
    \textbf{Model}      & \textbf{Sensors}  &  \textbf{Dataset} &  \textbf{mAP/AP} &  \textbf{FPS}   \\ \hline
     Faster R-CNN \cite{Ren2015FasterRCNN}   &  Camera   &   VOC2007  &  73.2         &  7             \\ 
     YOLO \cite{YOLO}   &  Camera           &   VOC2007  &  66.4         &  21             \\ 
     SSD  \cite{Liu2015SSDSS}  &  Camera    &   VOC2007  &  76.8         &  22             \\ \hline
     VoxelNet \cite{VoxelNet}  & LiDAR      &   KITTI           &  49.1        &  4.4            \\ 
     SECOND   \cite{SECOND}    & LiDAR      &   KITTI           &  56.7        &  20             \\
     PointPillar \cite{Lang2018PointPillarsFE}  & LiDAR  &  KITTI  & 59.2      & 62              \\ \hline
     F-PointNet \cite{FrustumPF}         &  Camera \& LiDAR &  KITTI    & 70.4   &  6      \\ 
     MV3D \cite{MV3D}    &   Camera \& LiDAR  & KITTI   &  62.35     &   2.78 \\ 
     AVOD \cite{AVOD}    &   Camera \& LiDAR &  KITTI    & 71.9   &  12.5      \\ \hline

    \end{tabular}
        \label{tab:object_detection}
    \vspace{-0.2 cm}
\end{table} 
\hfill \break
\textbf{Object Tracking:} Object tracking involves estimating object velocity and orientation, which can be utilized in the subsequent risk assessment and decision-making/planning systems. 
The integration of 3D LiDAR data points and 2D camera images is employed in \cite{Asvadi20163DOT}  to enhance the robustness of 3D object tracking. This system incorporates two parallel mean-shift algorithms for object detection and localization, respectively, and leverages a Kalman filter for the final fusion of previous information and tracking of detected objects. Likewise \cite{Hwang2016FastMO} uses both 2D and 3D information to segment the input data, including the camera image and the LiDAR point cloud. 
Subsequently, two object detectors are employed: Fast R-CNN \cite{Girshick2015FastR} for 2D object detection and the spin image method \cite{Johnson1997SpinImagesAR, Johnson1999UsingSI} for 3D detection. A segment matching-based method is adopted, in the end, to track the objects from consecutive frames. 
 
\textbf{Traffic Light Perception:} Locating the traffic light and identifying its status is essential for self-driving systems, ensuring they navigate the intersection safely and adhere to traffic rules. Typically, camera-based object detection models, such as YOLO \cite{YOLO, Redmon2018YOLOv3AI, Redmon2016YOLO9000BF, Bochkovskiy2020YOLOv4OS} and SSD \cite{Liu2015SSDSS}, can be adapted to address these traffic light recognition tasks \cite{Pavlitska2023TrafficLR}. Janahiraman et. al. \cite{Janahiraman2019TrafficLD} explored to adopt two models, SSD \cite{Liu2015SSDSS} and Faster R-CNN \cite{Ren2015FasterRCNN}, on traffic perception task and found that the later one performers better than the former. Wang et. al. \cite{Wang2021TrafficLD} proposed a detection and recognition model based on YOLOv4 \cite{Bochkovskiy2020YOLOv4OS}, incorporating a feature enhancement mechanism and bounding box uncertainty prediction mechanism to improve the perception accuracy.  Lin et. al. \cite{Lin2021ATF} also divided the traffic light perception module into two stages, detection, and classification. The detection models is based on Faster R-CNN \cite{Ren2015FasterRCNN} and the classification is based on VGG16 \cite{Simonyan2014VeryDC}. A detailed overview of various models for traffic light perception can be found in \cite{Pavlitska2023TrafficLR}. \bigskip

\begin{figure*}
    \centering
    \includegraphics[width=\textwidth]{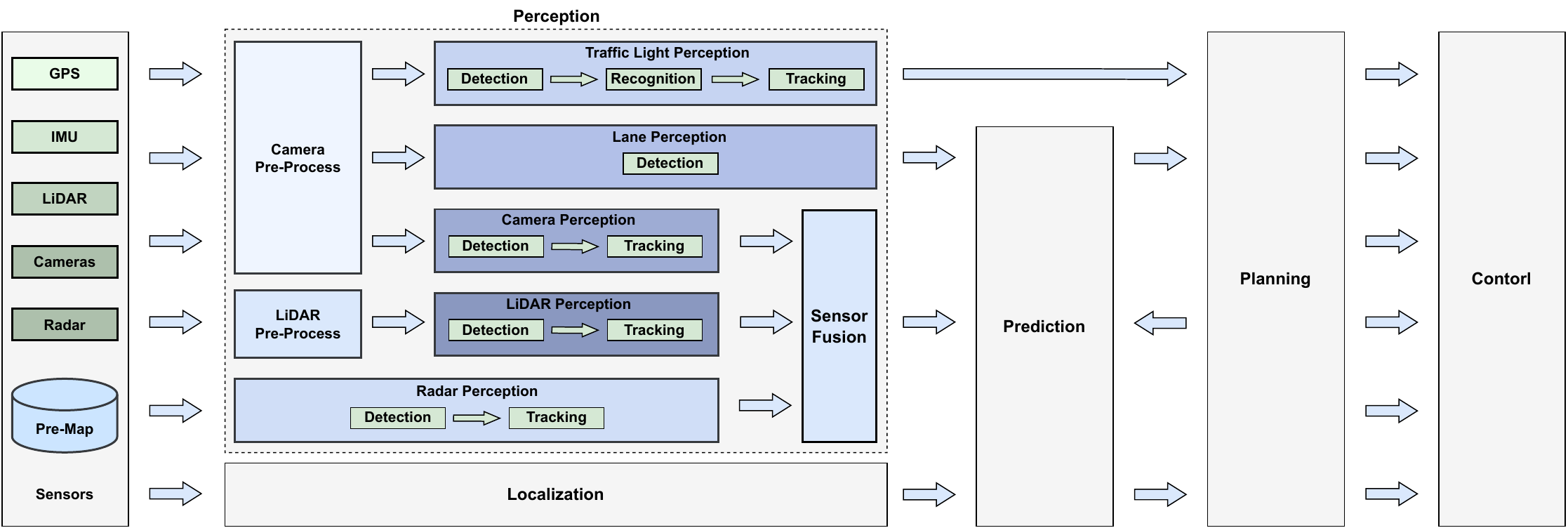}
    \caption{The Apollo system-driving framework based on version 9.0 with only perception breakdown for clarity.}
    \label{fig:Apollo}
\end{figure*}

\subsubsection{Localization} 
\hfill \break 
Determining the precise location of the ego-vehicle is crucial for autonomous driving, as the ego-vehicle must be aware of both its global and local positions. Global localization relies on GPS to determine latitude and longitude and corrects cumulative errors from the IMU, which measures angular rate and acceleration. Local localization uses sensors like LiDAR and cameras to detect road obstacles and shapes, generating occupancy maps.

A GPS and IMU fusion framework is adopted in \cite{Zhang2012ASF} to enhance localization precision, with cumulative errors mitigated by a mathematical algorithm. In \cite{Hata2014RoadMD}, an occupancy map with curbs and road markings is generated using an offline SLAM method, while online localization is achieved by detecting and matching these markers with LiDAR based on the offline map. In \cite{Suhr2017SensorFL}, instead of using high-cost LiDAR, a low-cost camera is employed to detect and classify road markings, and a low-complexity offline map is used to minimize computational costs. Additional localization methods are discussed in \cite{Kuutti2018ASO}.\bigskip

\subsubsection{Prediction}
\hfill \break
Prediction involves forecasting the future behaviors of road users, such as vehicles and pedestrians, by estimating their trajectories based on outputs from the perception and localization modules. Multiple-Trajectory Prediction (MTP) \cite{Cui2018MultimodalTP}, based on MobileNet, offers several possible trajectories for road actors, along with their associated probabilities.

Instead of using rasterized rendering, VectorNet \cite{Gao2020VectorNetEH} adopts a vectorized approach. It utilizes PointNet \cite{PointNet} to create polyline subgraphs and vectors, which are then integrated with graph neural networks (GNNs) to form a global graph for trajectory predictions. Additionally, VectorNet employs a BERT-like transformer architecture \cite{Devlin2019BERTPO} to capture global information on various trajectories and map polylines.

Wayformer \cite{Nayakanti2022WayformerMF} introduces a transformer-based model for fusing multimodal features related to roads, traffic lights, and road users, capturing both temporal and spatial information. It integrates data from various sources and explores different stages of input fusion. Wayformer employs an early fusion approach, combining multimodalities early in the processing pipeline due to its lower implementation complexity. Leveraging the results from perception and localization, the potential risk can also be estimated. A CNN-based overall risk assessment framework is proposed in \cite{Yurtsever2019RiskyAR}. It only requires the input from camera and evaluates the risk level for lane change scenarios.

Risk assessment at the software level should also be considered, as deep learning networks can sometimes lead to incorrect decisions that influence the final outcome. To address these uncertainties, a Bayesian Deep Learning (BDL) architecture is proposed in \cite{McAllister2017ConcretePF}. This approach allows the uncertainties of individual components to propagate through the architecture and be aggregated at the end, aiding in more informed and reasonable decision-making.\bigskip

\subsubsection{Planning}
\hfill \break 
With information from each system, a conclusive decision is essential for guiding and controlling vehicles. Planning (also known as decision-making) can be divided into global planning and local planning. Global planning involves mapping out the entire route to the destination, while local planning focuses on making immediate decisions based on the vehicle's current surroundings.

\textbf{Global Planning:} Due to the emergence of expressways and the increasing severity of traffic congestion, global planning has shifted its focus from finding the shortest route to the fastest route. Contraction Hierarchies (CH) algorithm \cite{Geisberger2012ExactRI} predicts the fastest route by including both the road hierarchies and road conditions. A road classification process is conducted first, automatically evaluating the assigned cost of each road segment, and node contraction algorithms \cite{Sanders2005HighwayHH, Goldberg2006ReachFA} filter out unimportant nodes while preserving the shortest routes. Then the fastest route is predicted by factoring the route conditions.

\textbf{Local Planning: } State lattice algorithm \cite{Pivtoraiko2005EfficientCP} discretizes continuous space into a state of lattice, transferring the motion planning problem to a graph search problem. Potential paths are constructed first by connecting the discrete states in the lattice, and a cost map with road conditions can be overlaid on it. Utilizing the search algorithms, such as \cite{Kushleyev2009TimeboundedLF} and \cite{Rufli2009OnTA}, the optimal path can be found.\bigskip

 \subsubsection{Apollo, a generic modular system example} 
\hfill \break
The Apollo autonomous driving system exemplifies the modular approach commonly adopted in self-driving applications. (refer to Fig.\ref{fig:Apollo} for a visual representation).

The perception module plays a pivotal role in identifying the environmental context. According to the ninth release \cite{Apollo}, this module encompasses obstacle perception, lane perception, and traffic light perception. Obstacle detection leverages data from cameras, LiDARs, and radar. A multi-task model, based upon YOLO \cite{Redmon2018YOLOv3AI, Redmon2016YOLO9000BF}, is employed to process and analyze visual data captured by cameras. For analyzing the 3D point cloud obtained from LiDAR, a CenterPoint \cite{Yin2020Centerbased3O} network is adapted. Additionally, a Kalman filter \cite{Kulathunga2019MultiCameraFI} can be employed to fuse the camera and LiDAR to comprehensively understand the surroundings. Traffic light perception initiates with proposing the region where the traffic light is likely located, followed by employing a model akin to Faster R-CNN \cite{Ren2015FasterRCNN} to determine the exact location. Then, a CNN model is applied to recognize the color. Lane detection can be performed by utilizing two models, denseline and darkSCNN \cite{Pan2017SpatialAD}.

With the information from the perception module, the prediction module estimates the behavior of obstacles using four sub-modules: container, scenes, evaluator, and predictor. The container aggregates the information from perception, localization, and planning modules while the evaluator utilizes this information to assess the path and speed of the objects. Ultimately, the predictor forecasts their trajectories using a composite network that incorporates a CNN, a MoblieNetv2 \cite{Sandler2018MobileNetV2IR}, a Long-short Term Memory (LSTM) network, and a multilayer perceptron (MLP) \cite{Xu2020DataDP}. The localization module determines the location of the ego-vehicle using data solely from only IMU and GPS or by integrating additional information from LiDAR and the camera. Building upon this, the planning module integrates the information from the aforementioned modules along with HD-Map to generate trajectory information, such as speed and acceleration, which are conveyed to the control module. Table \ref{tab:Baidu_Software} summarizes the ML models utilized in Apollo.

\begin{table}[ht]
    \centering
        \vspace{-0.2 cm}
    \caption{Subset of machine learning models implemented in Baidu Apollo }
        \vspace{-0.2 cm}
    \begin{tabular}{c|c|c}
    \hline
    \textbf{Model}      & \textbf{Type}  &  \textbf{Function}  \\ \hline
     YOLO \cite{Redmon2018YOLOv3AI, Redmon2016YOLO9000BF}  &  CNN   &  Object Detection              \\ 
     CenterPoint \cite{Yin2020Centerbased3O} &  CNN+FNN   &   Object Detection \\ 
     Faster-RCNN  \cite{Ren2015FasterRCNN}  &  CNN   &  Traffic Light Detection       \\
     darkSCNN \cite{Pan2017SpatialAD}  & CNN      &   Lane Detection                   \\ 
     MoblieNetv2 \cite{Sandler2018MobileNetV2IR}   &  CNN      &   Trajectories Prediction                    \\
     LSTM \cite{Xu2020DataDP} & RNN  & Trajectories Prediction              \\ 
     MLP  \cite{Xu2020DataDP} & FNN  &  Trajectories Prediction        \\ \hline

    \end{tabular}
        \label{tab:Baidu_Software}
    \vspace{-0.2 cm}
\end{table}

\subsection{End-to-end Software Systems}
\label{sec:end_to_end_systems}

Instead of separating the self-driving problem into a few sub-tasks and solving them individually, the end-to-end system integrates all components for a holistic solution. It takes the raw data from sensors as input and uses a single model to process the data and directly output the control commands. As the system is based on a single model, all functions can be trained together. Although the generic modular systems offer good interpretability and maintainability, the overall accuracy is highly dependent on the ability to avoid the errors from each module to propagate through the pipeline~\cite{McAllister2017ConcretePF}. The end-to-end system, however, mitigates this issue by using a single module, reducing the potential for error accumulation\cite{Chen2023EndtoendAD}. 

In general, the end-to-end system can be divided into two categories: imitation learning (IL) \cite{Kendall2018LearningTD, Liang2018CIRLCI, Chitta2022TransFuserIW, Shao2022SafetyEnhancedAD} and reinforcement learning (RL) \cite{Bojarski2016EndTE, Hawke2019UrbanDW, Codevilla2019ExploringTL}.\\

\subsubsection{Imitation Learning}
\hfill \break
Imitation learning is aimed at training the system based on human or expert actions such as steering and speed controls. Inspired by the ALVINN system \cite{Pomerleau2015ALVINNAA}, which is based on a tiny fully-connected network to achieve an end-to-end driving system, NVIDIA proposed an end-to-end system \cite{Bojarski2016EndTE} that combines and optimizes the individual tasks simultaneously, resulting in a smaller and better performance in contrast to the modular systems. This system was trained using only steering signals and was applied to an NVIDIA $\text{DRIVE}^{\text{TM}}$ PX computer platform with a 30 FPS. 

Besides leveraging the steering signals, the longitudinal control signals can also be taken into account \cite{Hawke2019UrbanDW}. The end-to-end system in \cite{Hawke2019UrbanDW} is divided into three components, perception with PWC-Net as backbone \cite{Sun2017PWCNetCF}, sensor fusion with a SAGAN \cite{Zhang2018SelfAttentionGA} for self-attention, and control. Similar to the previous system, CILRS \cite{Codevilla2019ExploringTL} also leverages both the steering and longitudinal signals as training inputs with ResNet34 \cite{He2015DeepRL} as a backbone to perceive the surrounding environment.

Imitation learning systems can also leverage a transformer architecture \cite{Vaswani2017AttentionIA}, aiming to capture long-range dependency in input data and enable parallel computation, to fuse different modalities. Different inputs are weighted by a multi-head self-attention mechanism. According to the weighted results, the architecture focuses on a specific task.

\begin{figure}[ht]
    \vspace{-0.2cm}
    \centering \includegraphics[width=0.485\textwidth]{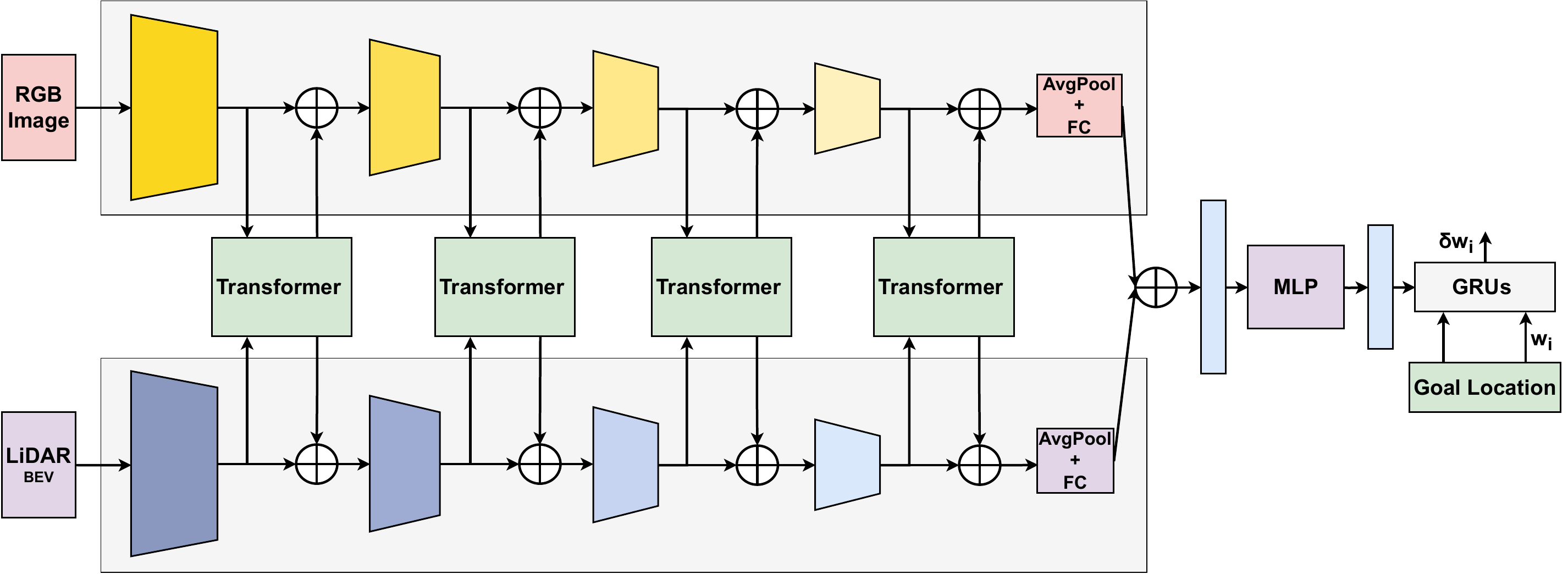}
    \caption{An overview of transformer-based TransFuser architecture.}
    \label{fig:TransFuser}
    \vspace{-0.2cm}
\end{figure}

TransFuser \cite{Chitta2022TransFuserIW}, illustrated in Fig. \ref{fig:TransFuser}, takes the data from both the camera and its typical complementary sensor, LiDAR, to improve the performance of self-driving cars in complex driving scenarios. It utilizes ResNets \cite{He2015DeepRL} and RegNets \cite{Cardoso2016AMM} as the backbone architecture and adopts several transformer modules to fuse different resolution feature maps. An element-wise summation is used to combine the outputs. To address the complex driving environment, different tasks, such as waypoint prediction, HD map prediction, and vehicle detection, etc, are implemented.

\begin{figure}[ht]
    \vspace{-0.2cm}
    \centering \includegraphics[width=0.485\textwidth]{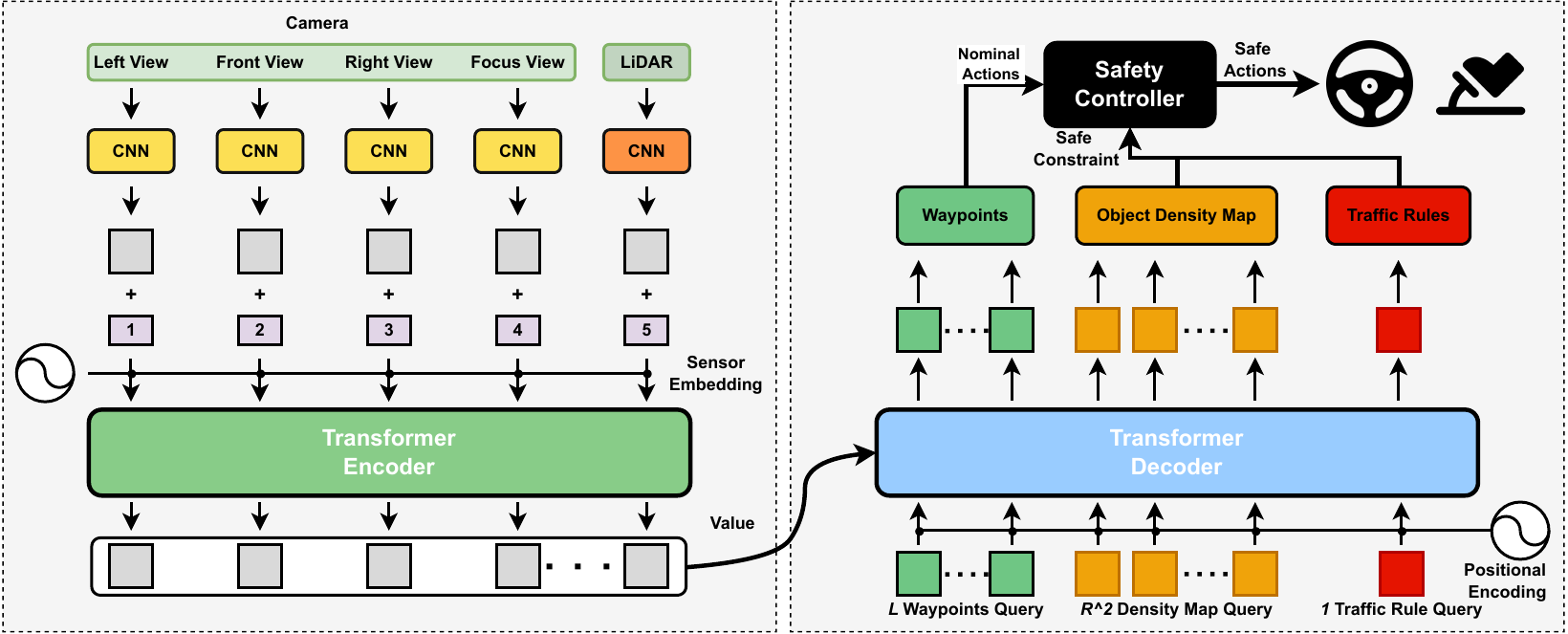}
    \caption{An overview of transformer-based InterFuser architecture used in \cite{Shao2022SafetyEnhancedAD}.}
    \label{fig:InterFuser}
    \vspace{-0.2cm}
\end{figure}

InterFuser \cite{Shao2022SafetyEnhancedAD}, illustrated in Fig.\ref{fig:InterFuser}, focuses on improving the safety of self-driving by generating safety mind maps, also called intermediate interpretable features, with information about the surrounding environment. Similar to TransFuser, it processes data both from the camera and LiDAR. However, multi-view features from different cameras are leveraged to achieve a comprehensive understanding of the surroundings. As shown in Fig.\ref{fig:InterFuser}, a transformer network is adopted to weigh the features extracted by CNN backbones, ResNets \cite{He2015DeepRL}, from multiple views of multi-sensors. Leveraging the output from the transformer architecture, waypoints prediction, object density maps, and traffic rules functions can be implemented to improve the safety and efficiency of self-driving.

\begin{figure*}
    \centering
    \includegraphics[width=\textwidth]{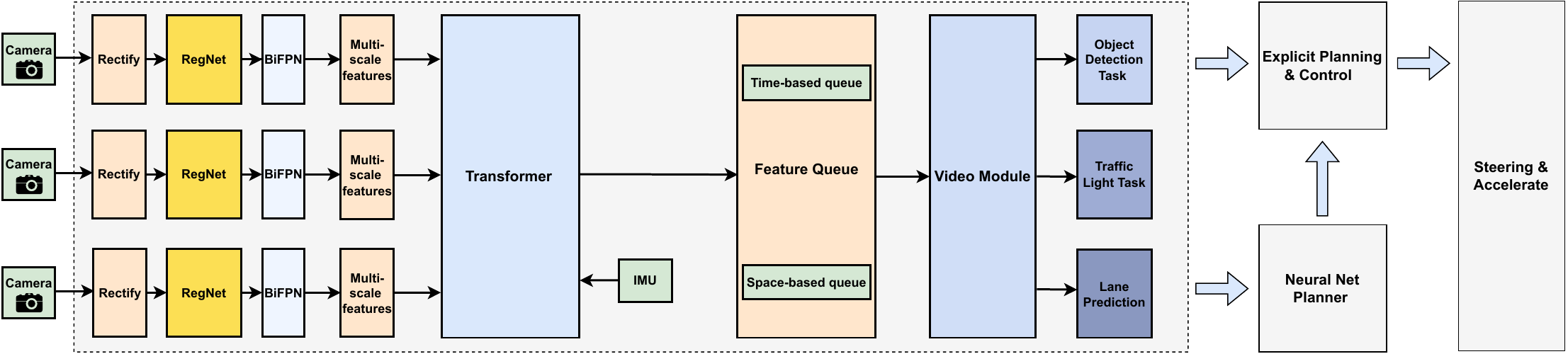}
    \caption{Simplified system diagram of Tesla's autonomous system.}
    \label{fig:Tesla}
\end{figure*}
\bigskip

\subsubsection{Reinforcement Learning}
The reinforcement learning system aims to treat self-driving as a Markov decision problem \cite{Silver2016MasteringTG} where the agent, in this case self-driving car, interacts with the environment and tries to maximize its cumulative reward. Wayve demonstrates the first application of reinforcement learning in the self-driving car \cite{Kendall2018LearningTD}. It is based on the Deep DPG (DDPG) network \cite{Lillicrap2015ContinuousCW} and only requires a monocular image as input to achieve lane following. The automated traveling distance without human inference is set as a reward. CIRL \cite{Liang2018CIRLCI} integrates a controllable imitation stage and a reinforcement learning stage. The imitation stage trains the network with human actions and shares the result weights with the DDPG-based reinforcement learning stage to solve the local optimal issues.\bigskip
\subsubsection{Large Language Models}
A recent research explores integrating object-level vector data with Large Language Models (LLMs) for autonomous driving \cite{chen2024driving}. This multimodal fusion approach enhances the interpretability of decisions made by reinforcement learning models, enabling autonomous systems to better process driving scenarios with explainable reasoning.\bigskip
\subsection{Tesla: example of end-to-end system}
Fig. \ref{fig:Tesla} presents a simplified software system diagram derived from 2021 Tesla AI day \cite{TeslaAI2021}. As an example of the end-to-end system, Tesla employs rectification layers to correct image distortions, followed by RegNets \cite{Radosavovic2020DesigningND} employed to process the images into different scales and resolution features. These are then linked via a Bidirectional Feature Pyramid Network (BiFPN) \cite{Tan2019EfficientDetSA} to facilitate inter-layer information sharing. A transformer model \cite{Vaswani2017AttentionIA} with a self-attention mechanism is implemented to weigh the importance of different input features, resulting in a more efficient perception from different cameras. Subsequently, the feature queue module caches the features, including the position, multi-camera features, and ego kinematics, and then concatenates them. Two types of queues are included: a time-based queue to handle occlusion and a space-based queue for road geometry prediction. Then a video module leverages a spatial RNN to fuse these cached frames temporally and only update the nearby points with RNNs. With the online map generated by the video module, different tasks can be employed, including object detection and lane prediction. Finally, a neural net planner predicts the optimal trajectory leveraging the vector space generated by the vision module based on a Monte Carlo tree search algorithm.

\section{Hardware Systems}
\label{sec:hardware_section}
\begin{figure*}[h!]
\centering
\begin{minipage}{.2\textwidth}  
  \centering
  \subfloat[]{\label{fig:CPU+GPU}\includegraphics[width=\textwidth]{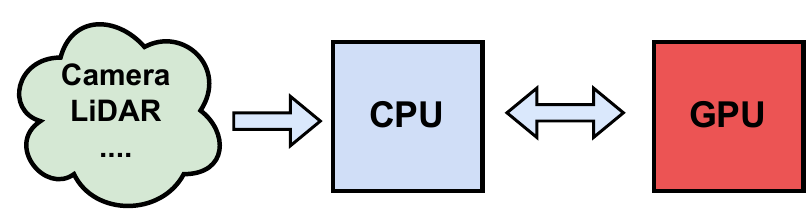}}
  \vfill
  \subfloat[]{\label{fig:FPGA+GPU}\includegraphics[width=\textwidth]{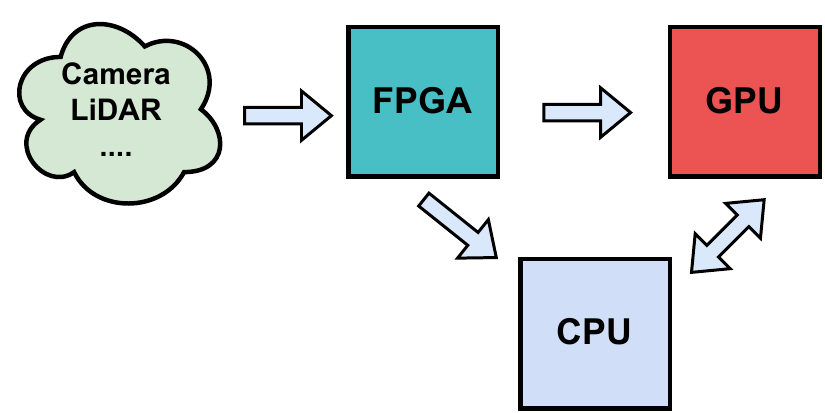}}
\end{minipage}
\hspace{0.05\textwidth}  
\begin{minipage}{.25\textwidth}  
  \centering
  \subfloat[]{\label{fig:NNA+GPU}\includegraphics[width=\textwidth]{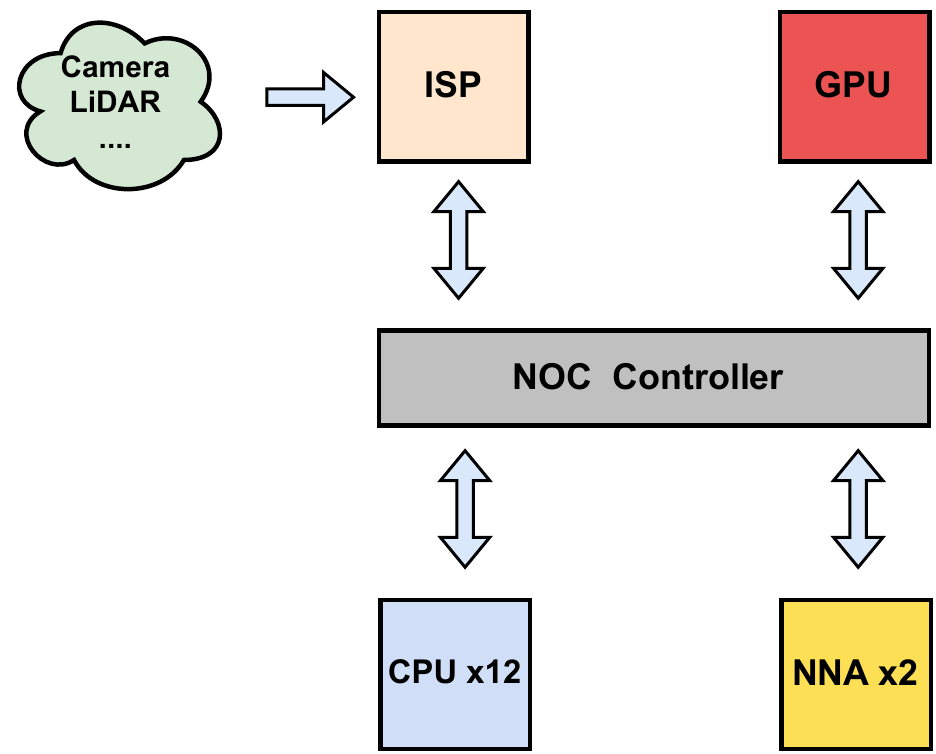}}
\end{minipage}

\caption{(a) Conventional CPU+GPU-based system, (b) Integrating an FPGA as a data processor on CPU+GPU-based system adopted from \cite{NVIDIA2024PonyAV}, (c) Integrating dedicated neural network accelerators (NNA) in the system, adopted from \cite{Talpes2020ComputeSF}.}
\vspace{-0.3 cm}
\end{figure*}

Short inference time and low power consumption are critical for self-driving cars. The substantial computational demands of these tasks often require more than just conventional Central Processing Units (CPUs), as CPUs are not inherently designed for massively parallel computing. In the following subsections, we discuss the most commonly used hardware in autonomous driving systems, which typically work in tandem with CPUs to meet these demands.\\

\subsection{Graphics Processing Units (GPUs)}
Graphics Processing Units (GPUs) are essential in accelerating self-driving systems by leveraging their ability to perform parallel computations. Unlike traditional CPUs, which handle tasks sequentially, GPUs can process multiple tasks simultaneously, significantly enhancing the speed and efficiency of complex computations required for autonomous driving. This parallel processing capability has made GPUs a preferred choice in modern self-driving systems, including those used by Waymo [5] and Apollo [6]. For example, the Apollo platform recommends two types of industrial-grade GPU computers: the Nuvo-6108GC and the Nuvo-8108GC. The Nuvo-6108GC can integrate an NVIDIA RTX 3070 GPU with an Intel Xeon processor, offering a computational capacity of up to 20.31 TFLOPS. The Nuvo-8108GC supports GPUs with a maximum computational capacity of 82.6 TFLOPS, taking advantage of a higher power budget. 

In the initial phase of Pony.ai \cite{PonyAI2024}, a CPU+GPU setup was employed, where the CPU managed sensor data and scheduled computational tasks, while the GPU handled machine learning inference, as shown in Figure  \ref{fig:CPU+GPU}). However, this configuration placed significant demands on the CPU. The addition of more sensors or an increase in input resolution can substantially raise the computational load, potentially overwhelming the CPU and creating a system bottleneck. This bottleneck can degrade overall performance, affecting critical metrics such as FPS, and limit the deployment of more sophisticated machine learning models.\\
While GPUs accelerate inference through parallel computation, this process demands significant power, leading to substantial heat generation. This heat presents a thermal management challenge in vehicles, which must be carefully addressed to maintain system performance and reliability.\cite{Lin2018TheAI}.

\begin{table*}[ht]
    \centering
        \caption{The summary of hardware platforms for modern self-driving systems}
    \begin{tabular}{c|c|c|c|c|c}
    \hline
    \textbf{Components} & \textbf{Nuvo-6108GC \cite{neousys_nuvo6108gc}}  & \textbf{Jetson AGX Orin 32GB \cite{nvidia_jetson_orin}}& \textbf{DRIVE Orin SoC \cite{NVIDIA_DRIVE_ORIN}} & \textbf{FSD SoC \cite{Talpes2020ComputeSF}} & \textbf{EyeQ\texttrademark~6 High \cite{mobileye_eyeq}} \\ \hline
    \textbf{CPU} &  Intel Xeon or Core & Cortex-A78AE \texttimes 8  &  Cortex-A78A \texttimes 12    & Cortex-A72 \texttimes 12&  CPU Cores \texttimes 8     \\ 
    \textbf{GPU} &  RTX 3070 GPU & Ampere GPU  & Ampere GPU                   & Mali-G71                    &    Dedicated      \\ 
    \textbf{ISP} &  - & \checkmark & \checkmark                    & \checkmark                  & \checkmark \\
    \textbf{DLA} & - & Gen2 DLA \texttimes 2   & Gen2 DLA \texttimes 2        & NNA \texttimes 2            & XNN6 \texttimes 4  \\ 
    \textbf{VLIW\&SIMD}& - & Gen2 PVA  & Gen2 PVA   & SIMD within NNA  & VMP6 \texttimes 4\\ 
    \textbf{CGRA} & - & - &   -  & - & PMA6 \texttimes 2 \\ 
    \textbf{TOPS} & 162.6 & 200  &  254 &  72 & 34 \\ 
    \textbf{TDP (W)} & 220 & 40 &  200  & 40 & 33 \\ 
    \textbf{Brand} & Neousys & NVIDIA & NVIDIA  & Tesla & Mobileye \\
    \textbf{Deployment} & Apollo, etc. &  Apollo, etc. & Volvo Cars, etc. & Tesla & GEELY, etc.\\\hline
     \end{tabular}
    \vspace{0.2cm}

    \label{tab:case-hardware}
    \vspace{-0.5 cm}
\end{table*}

\subsection{Field-Programable Gate Arrays (FPGAs)}
To address the aforementioned issues with GPUs, an FPGA can be used to manage sensor data processing \cite{NVIDIA2024PonyAV} or serve as a more energy-efficient hardware accelerator \cite{Sciangula2022HardwareAO}. The inherent reconfigurability of FPGAs offers significant advantages, especially in the early development phases of autonomous driving systems. For instance, Pony.ai \cite{PonyAI2024} utilized an FPGA to handle sensor data, reducing the load on the CPU. To minimize latency in data transmission, the FPGA could directly transfer the necessary data to the GPU, bypassing the CPU, as illustrated in Fig. \ref{fig:FPGA+GPU}.

Another example of using FPGAs in autonomous driving is based on the Apollo perception framework, where an FPGA was demonstrated as an energy-efficient alternative to conventional GPUs \cite{Sciangula2022HardwareAO}. This study focused on accelerating various neural networks within the perception framework, highlighting both the advantages and limitations of this approach.

The implementation was carried out on the embedded Deep Learning Processor Unit (DPU) core in a Zynq Ultrascale+ MPSoc ZCU102 FPGA SoC platform. This platform integrates a reconfigurable FPGA unit and a multi-core processing processor. The processing unit acts as a host, processing software tasks and overseeing hardware tasks executed on the reconfigurable unit. 
The neural networks were partitioned into two parts: one part executed on FPGA to explore its parallel processing capabilities, while the other managed the remaining computation tasks less suited for FPGA. The results indicated that the FPGA accelerator outperformed the GPU in several tasks, including the denseline lane tracker and traffic light detection.

\subsection{Systems-on-Chip (SoCs)}
\label{subsec:SoCs}
With the increasing demand for computational efficiency and reduced power consumption, autonomous driving systems are increasingly adopting System-on-Chip (SoC) architectures. SoCs provide an integrated solution that boosts performance while minimizing power usage and latency, making them ideal for the complex tasks required in self-driving systems.\\
Tesla has designed the world’s first purpose-built full self-driving (FSD) computer \cite{Talpes2020ComputeSF}. The core of this system features two FSD chips, each built with a System-on-Chip (SoC) architecture that integrates a GPU, twelve CPUs, two dedicated neural network accelerators (NNAs), and an image signal processor (ISP), as illustrated in Fig.\ref{fig:NNA+GPU}. The ISP handles the preliminary processing of image data, and then transfers it to the DRAM. The CPUs coordinate tasks between the NNAs and the GPU, directing each to manage specific algorithms, with the GPU also responsible for post-processing duties. Leveraging these integrated components, one single FSD chip can provide 72 TOPS computation capacity with maximum power consumption below 40 W.

NVIDIA has released the DRIVE Orin \cite{NVIDIA_DRIVE_ORIN}  platform, an advanced autonomous vehicle computer. Like the FSD chip, the DRIVE Orin is based on SoC architecture. This system includes an Ampere architecture GPU, 12 ARM Cortex-A78 CPUs, two deep learning accelerators (DLAs), a programmable vision accelerator (PVA), and an optical flow accelerator (OFA). Together, these components enable the DRIVE Orin to deliver a total of 254 TOPS with a maximum power consumption of 200 W, as reported for its developer kit.

Similar to the DRIVE Orin, the Jetson Orin \cite{nvidia_jetson_orin} is also based on the Orin SoC architecture but features distinct configurations, as detailed in Table \ref{tab:case-hardware}. Beyond industrial edge AI platforms that utilize commercial GeForce GPUs, Apollo also supports systems equipped with Jetson Orin SoCs. These include the 32GB and 64GB AGX Orin series, as well as the Orin NX 16GB variant. The 32GB Jetson AGX Orin delivers up to 200 TOPS, while the 64GB version offers up to 275 TOPS. Although the Jetson Orin NX 16GB variant provides lower computational capacity, it is highly power-efficient, with a maximum power consumption of 25 W.

Mobileye has launched the EyeQ\texttrademark series \cite{mobileye_eyeq}, a line of SoC chips tailored for autonomous driving. These processors are engineered for heterogeneous computing and scalable architecture, integrating various compute density computation units, from general-purpose CPUs to function-specific accelerators. EyeQ\texttrademark~6 High contains an eight-core CPU cluster, an ISP, a GPU, two general-compute accelerator clusters, and two dedicated DLA clusters. Each general-compute accelerator cluster encompasses one programmable macro array (PMA), two vector microcode processors (VMPs), and two multi-threaded processor clusters (MPCs). The PMA is a CGRA-like processing array, optimizing for tasks in computer vision and deep learning, while VMPs handle short-integer computations with a combination of Very-Large Instruction World (VLIW) and Single Instruction Multiple Data (SIMD) architectures. The MPCs can tackle multiple threads simultaneously, allowing for efficient processing of varied computational tasks. Each DLA cluster in the EyeQ\texttrademark~6 has two XNN engines to accelerate the convolution neural networks (CNNs). Consequently, with these components, the whole system can offer 34 TOPS with a maximum power consumption of 33 W.

\section{Moving Forward: Hardware Accelerators}

\subsection{Scaling Current Hardware}

The demand for real-time perception and control in autonomous driving systems, coupled with the shift toward higher levels of autonomy (Levels 4 and 5), requires more powerful and capable hardware. As these higher levels of autonomy involve increasingly complex modules and models across the system's full stack, along with a greater number of inputs from various sensors, robust processing capabilities become essential.
Based on Table \ref{tab:case_study}, most of our studied self-driving systems except for Tesla,  commonly utilize a hybrid system of CPUs and GPUs to tackle these high computation demands. \\

\begin{table}[ht]
    \centering
        \vspace{-0.2 cm}
        \caption{Hardware configuration of self-driving cars}
    \begin{tabular}{c|c|c|c}
    \hline
    \textbf{Model}  &  \textbf{CPU} & \textbf{GPU} & \textbf{NNA}  \\ \hline
     IARA \cite{Badue2019SelfDrivingCA}     & \checkmark   & \checkmark & -         \\ 
     Waymo \cite{Waymo}     & \checkmark   & \checkmark & -           \\ 
     Tesla HW3 \cite{Tesla} & \checkmark    & \checkmark & \checkmark         \\ 
     Tesla HW4 & \checkmark    & - & \checkmark     \\      
     Apollo \cite{Apollo}    & \checkmark    &  \checkmark & -    \\ \hline
     \end{tabular}

    \label{tab:case_study}
    \vspace{-0.2 cm}
\end{table}

To evaluate the performance of different self-driving systems across various off-the-shelf hardware setups, we selected three end-to-end benchmarks: TransFuser \cite{Chitta2022TransFuserIW}, InterFuser \cite{Shao2022SafetyEnhancedAD}, and MILE  \cite{Hu2022ModelBasedIL} (Table \ref{tab:Char_three}). The specifications of the hardware system used for these evaluations are also summarized in Table\ref{tab:GPU_SPEC}. As shown in the table, the CPUs and GPUs provide a wide range of capabilities, representing a broad spectrum of computational power.

\begin{table}[ht]
    \centering
        \vspace{-0.2 cm}
        \caption{Characterization of three end-to-end self-driving models used for evaluation}
    \begin{tabular}{c|c|c|c}
    \hline
    \textbf{Model}  &  \textbf{Trainable Param. (M)} & \textbf{GFLOPs} & \textbf{FLOPs/Byte}  \\ \hline
     TransFuser \cite{Chitta2022TransFuserIW}     & 165.59   & 67.45 & 101.83        \\ 
     InterFuser \cite{Shao2022SafetyEnhancedAD}     & 48.09  & 32.64 & 169.68       \\ 
     MILE \cite{Hu2022ModelBasedIL} & 88.29   & 54.04 & 153.02         \\ \hline
     \end{tabular}

    \label{tab:Char_three}
    \vspace{-0.2 cm}
\end{table}

\begin{table*}[ht]
    \centering
        \caption{Configuration of systems used for benchmark evaluation}
    \begin{tabular}{c|c|c|c|c|c|c|c}
    \hline
    \textbf{System} & \textbf{CPU} & \textbf{Threads} & \textbf{Freq (GHz)} & \textbf{PCIe} & \textbf{GPU} & \textbf{CUDA Cores}  &  \textbf{TDP (W)}\\ \hline
     1 & Cortex-A78AE     &  12 & 2.2 & Gen4 & Jeston AGX Orin  & 2048   &  60\\ 
     2 & Ryzen 9 5900HX   & 16 & 4.6 & Gen3  & GeForce RTX 3060  & 3840   & 115 \\ 
     3 & Core i9-12900K & 24 & 5.2 & Gen5 & RTX A4000 &  6144  & 140 \\ 
     4 & Ryzen 9 7950X    & 32 & 5.7 & Gen5 & GeForce RTX 4090 & 16384  &  450  \\  \hline    
     \end{tabular}

    \label{tab:GPU_SPEC}
\end{table*}

\begin{figure*}[htbp]
\centering
   \begin{minipage}{\textwidth}
   
   \begin{minipage}{0.5\textwidth}
   \centering
   \includegraphics[width=\textwidth]{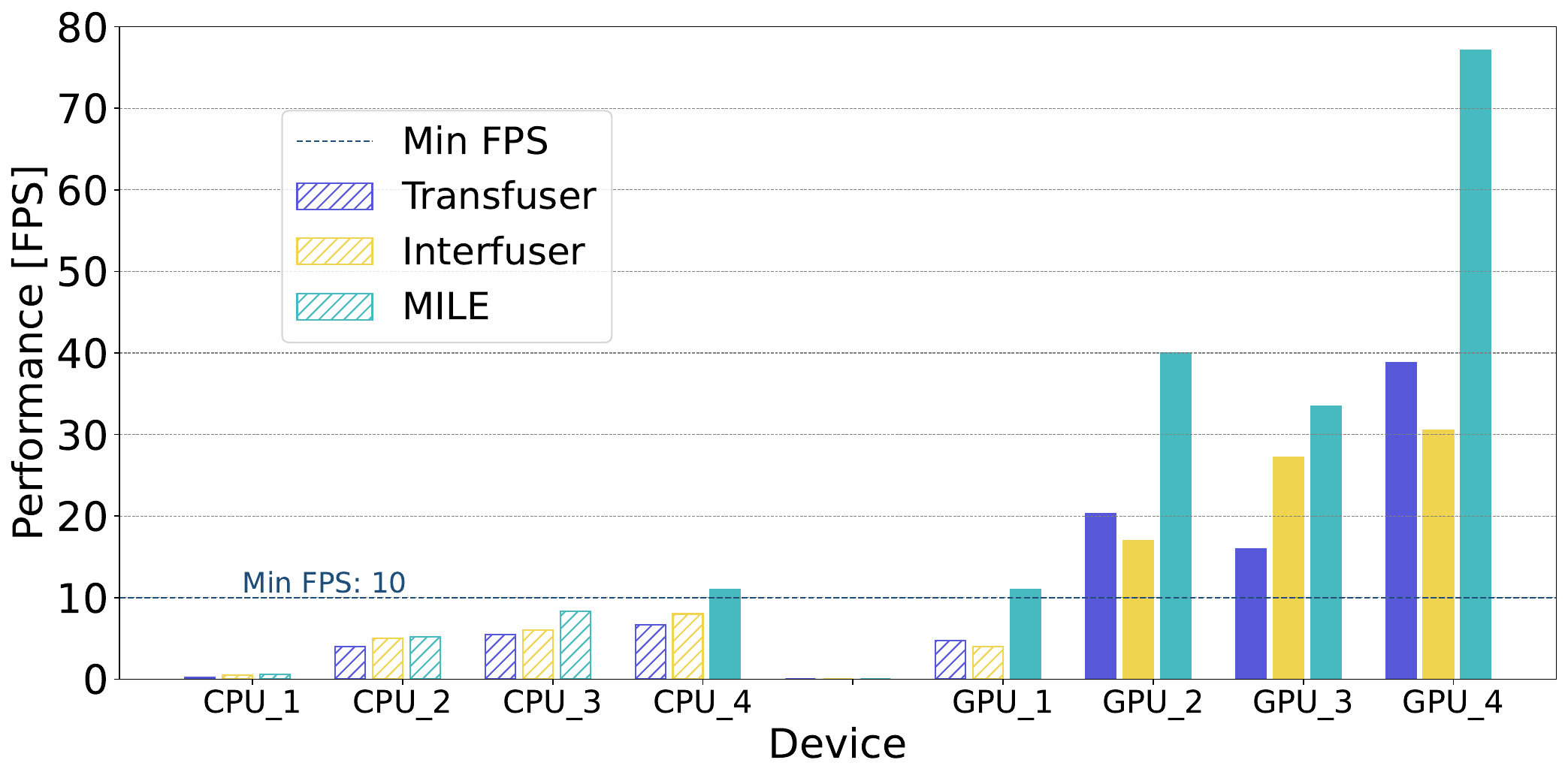}
   {\small (a)}
   \end{minipage}
   \hspace{0.01\textwidth}
   \begin{minipage}{0.5\textwidth}
   \centering
   \includegraphics[width=\textwidth]{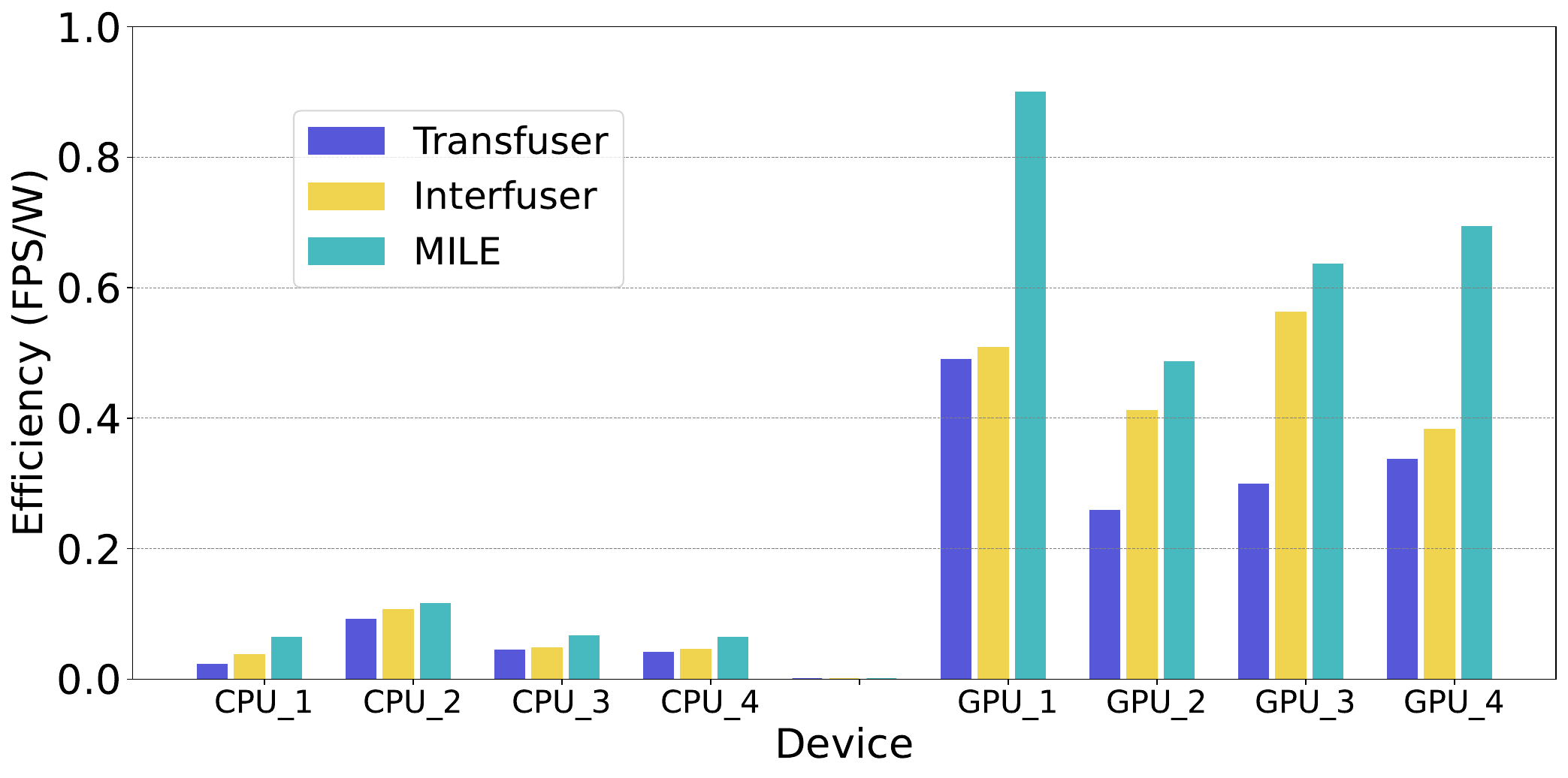}
   {\small (b)}
   \end{minipage}
       
   \end{minipage}

\caption{(a) Performance (in terms of Frame Per Seconds) of the three end-to-end systems on various devices  (b) Efficiency (in terms of performance per watt) of these various devices. Hatched bar indicate  performance below the required level.}
\label{fig:profiling_result}
\end{figure*}

Fig.\ref{fig:profiling_result} illustrates the performance and efficiency of the three evaluated benchmarks on the specified device of the hardware systems mentioned in Table \ref{tab:GPU_SPEC}. As shown in Fig.\ref{fig:profiling_result}(a), none of the CPUs solely —except for one with borderline results— can meet the minimum required performance of 10 FPS for self-driving systems. This finding aligns with the industry's shift toward using GPUs as the primary accelerators for these workloads. As expected, GPUs deliver significantly better performance than CPUs, especially in more capable devices.
In Fig.\ref{fig:profiling_result}(b), although GPUs generally have a higher Thermal Design Power (TDP), they also demonstrate superior efficiency. This is because their enhanced computational capabilities reduce processing time so substantially that the increased power consumption is offset, resulting in lower overall energy usage. Worth noting is that these results indicate even a single GPU can exhibit varying performance and efficiency across different models, emphasizing the need for more tailored solutions to achieve higher efficiency while maintaining performance above the required threshold.
This observation also supports the urgent need to radically reduce the carbon emissions from computation in autonomous vehicles (AVs), especially given projections that AVs will dominate 95\% of the market by 2050 \cite{9942310}.

\subsection{Tailored-to-task accelerators}
As discussed in Section \ref{sec:software_systems}, we are dealing with a multi-model heterogeneous system where a variety of fundamentally different tasks must be processed within a limited time frame. Managing these diverse models with a single type of hardware leads to suboptimal performance, as briefly demonstrated in the previous section. Given these challenges, significant efforts have been made in both research and industry to bridge this gap. For example, both Tesla and Mobileye have developed specialized, task-specific cores designed to handle different tasks across the full stack of software systems (see Subsection.\ref{subsec:SoCs})).
Sparsity is another key example of heterogeneity that needs to be captured. For instance, 3D point clouds derived from LiDAR exhibit high levels of sparsity, which presents challenges for conventional GPUs typically optimized for dense matrix operations.

\begin{figure*}[htbp]
\centering
   \begin{minipage}{\linewidth}

   \begin{minipage}{0.48\textwidth}
   \centering
   \includegraphics[width=\textwidth]{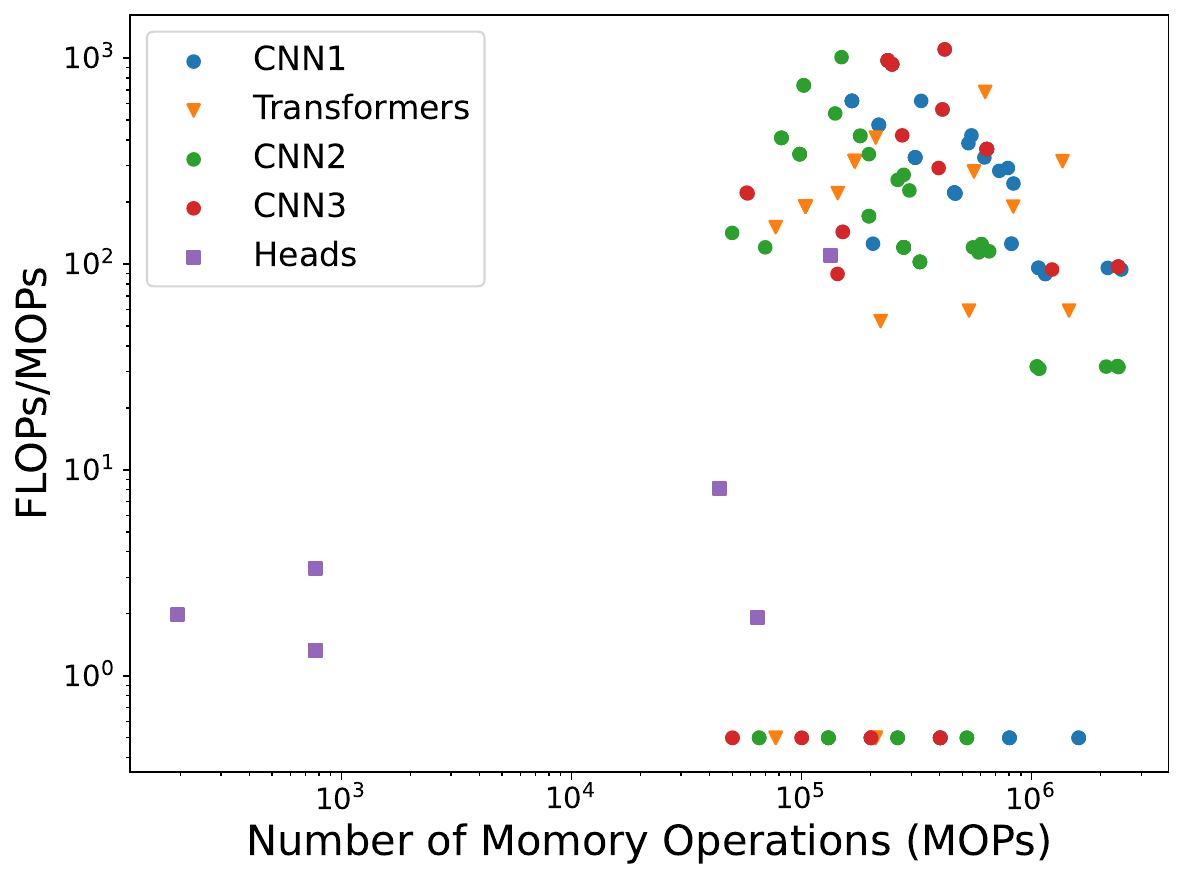}
   {\small (a)}
   \end{minipage}
   \hspace{0.01\textwidth}
   \begin{minipage}{0.49\textwidth}
   \centering
   \includegraphics[width=\textwidth]{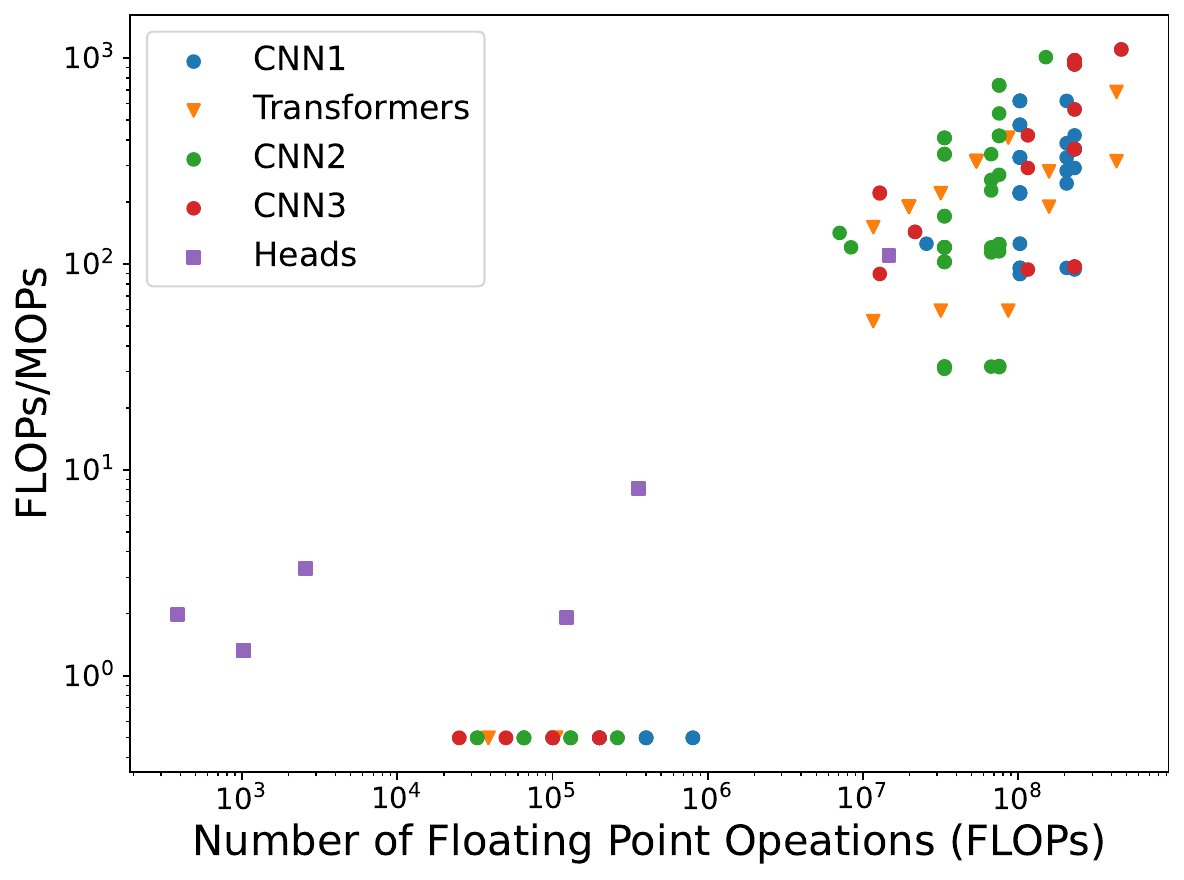}
   {\small (b)}
   \end{minipage}
       
   \end{minipage}

\caption{(a) The arithmetic intensity of sample layers in the InterFuser AD system versus memory operations (reading data from memory)  (b) The arithmetic intensity of sample layers in the InterFuser AD system versus number of floating point operations.}
\label{fig:arithmetic_intensity}
\end{figure*}

Beyond this coarse-grained heterogeneity, there is also significant variation among the different layers within each model used in autonomous driving systems. For instance, Fig.\ref{fig:arithmetic_intensity} illustrates the arithmetic intensity of sample layers in the InterFuser model. Similar to the heterogeneity observed between different tasks, achieving optimal performance and efficiency requires careful consideration of this layer-specific variation when designing accelerators. A critical consideration is striking the right balance between specialization and generality. Given the long lifespans of autonomous vehicles, hardware must retain some ability to adapt to future workloads, ensuring that it remains effective as new demands and models emerge.

\subsection{Processing-In-Memory: a promising solution}

While efficient dataflow can reduce data movement and improve energy efficiency, the bandwidth between the accelerator and off-chip memory, typically DRAM, often remains a bottleneck. This issue is particularly pronounced in multi-core systems, where a high degree of parallelism requires the memory to service multiple cores simultaneously. Additionally, data movement between off-chip memory and on-chip components is costly in terms of both energy consumption and latency \cite{Boroumand2017LazyPIMAE}.

Given these challenges, a potential solution is to integrate computation directly within or near the memory where data is stored—a concept known as processing-in-memory (PIM). This idea, originally proposed around fifty years ago \cite{Kautz1969CellularLA, Stone1970ALC}, involves embedding computation logic within memory itself. However, the adoption of PIM was initially limited by the lack of suitable technologies. Today advances in 3D-stacked memory technology and the development of new memory standards, such as High-Bandwidth Memory (HBM) and Hybrid Memory Cube (HMC), have revived interest in these ideas.

3D-stacked memory can provide higher bandwidth than conventional DDR memory, an HBM from Samsung can provide 1 TB/s bandwidth compared to only 32 GB/s for GDDR5 \cite{HBM2vsDDR5}. 
Besides putting the 3D-stacked memory off-chip, the accelerator can be integrated into its logic layer, fully utilizing the high internal bandwidth through the short "wires", vias. 

In the context of autonomous driving systems, where DNNs and CNNs are extensively used for tasks such as object detection, sensor fusion, and decision-making, PIM offers a compelling advantage to meet the stringent performance and efficiency demands of AD systems.

Neurocube \cite{Neurocube} is an example of a near-memory processing accelerator that integrates processing elements (PEs) within the logic die of the HMC, leveraging high internal bandwidth to accelerate neural network computations. Programmable neuro-sequence generators enable fully data-driven computing, with each vault connected to a PE and PEs communicating via a 2D mesh network. Each PE includes multiple MACs, cache memory, a temporal buffer, and a memory module to optimize data reuse. Synthesized using a 15-nm FinFet process, Neurocube achieves up to 132 GOPS for ConvNN \cite{Gould2009DecomposingAS} inference with a power consumption of 3.41 Watts.

\begin{figure}[htbp]
    \vspace{-0.2cm}
    \centering
    \includegraphics[width=0.48\textwidth]{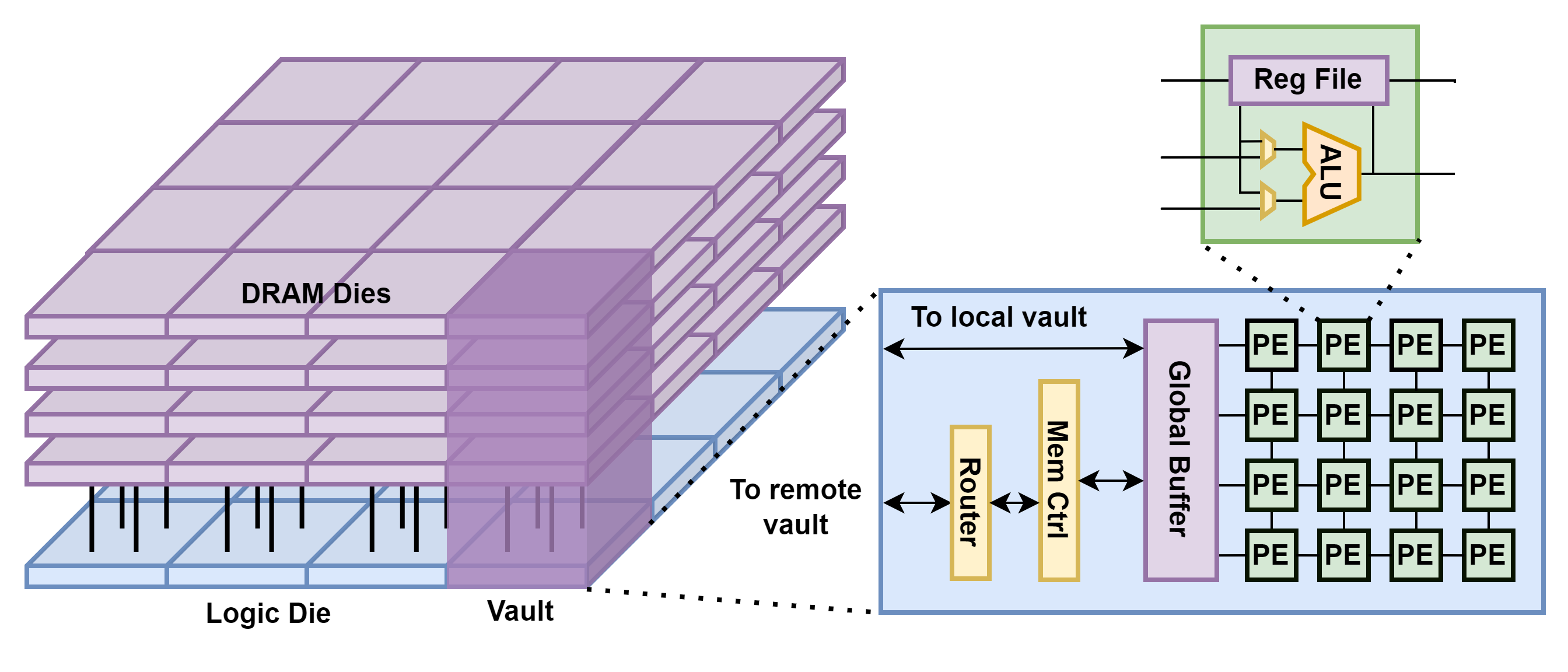}
    \caption{Architecture of TETRIS accelerator adopted from \cite{Gao2017TETRISSA}, actual array size is $14\times14$.}
    \label{fig:TETRIS}
    \vspace{-0.1cm}
\end{figure}

TETRIS \cite{Gao2017TETRISSA} is an HMC-based accelerator designed to enhance DNN performance by integrating the computation array directly into the memory’s logic layer (see Figure \ref{fig:TETRIS}). The memory stack is divided into sixteen vaults, each containing DRAM banks and a controlling logic. The DNN accelerator replaces the crossbar network in the controlling logic with a 2D mesh NoC, similar to the Eyeriss architecture \cite{Chen2016EyerissAE}. Each processing element (PE) in the array has a small register file and a fixed-point ALU, while a global buffer shared by all PEs supports data reuse. TETRIS achieves a 4.1x performance improvement and 1.5x energy savings compared to a 2D NN accelerator with 1024 PEs and four low-power DRAM channels. By making minor modifications to the memory cells of DRAM, simple computations can be implemented directly within the memory layers. This approach falls under the concept of using-memory-processing, a specific subcategory of processing-in-memory (PIM). 

Ambit \cite{Seshadri2017AmbitIA} supports simple bitwise operations within DRAM by slightly changing the control logic. By simultaneously enabling three rows, bitwise AND and OR operations can be performed. In addition, NOT operation can be achieved by modifying the sense amplifier. The capability to support basic bitwise operation provides the possibility to perform more complex operations, such as XOR operation and even addition and multiplication. SIMDRAM \cite{Hajinazar2021SIMDRAMAE} provides the ability to more efficiently performing complex operations, such as multiplication and addition, compared to Ambit \cite{Seshadri2017AmbitIA}. Complex operations can be decoupled to the majority (MAJ) and NOT operations to improve flexibility by supporting arbitrary operations. To facilitate the implementation of complex operations, a new programming interface, instruction set architecture (ISA) support, and hardware components are integrated. DrAcc \cite{Deng2018DrAccAD}, based on Ambit\cite{Seshadri2017AmbitIA}, incorporates a carry look-ahead adder within the DRAM to accelerate ternary weight neural networks (TWNs). It leverages the basic bitwise operation, AND, OR, and NOT to perform the addition operation. Several circuit modifications are performed to support carry shift and propagation. By utilizing the presented shift and addition circuit, multiplications required by TWN can also be implemented.

\subsection{The Road Ahead of Self-Driving Accelerators}
The evolution of autonomous driving systems will require increasingly sophisticated hardware platforms to meet the growing demands for real-time processing, energy efficiency, and adaptability. As AD software continues to evolve, driven by advances in deep learning models, sensor fusion, and decision-making algorithms, the underlying hardware must also keep the pace. While specialized cores tailored to specific tasks—such as image processing, sensor fusion, or neural network inference are essential for achieving optimal performance and high efficiency, there is also a need for flexibility. Given the 10- to 15-year lifespan of autonomous vehicles, the hardware must be capable of adapting to new algorithms, models, and software updates that will emerge over time. 
Based on these facts, we anticipate that the future of self-driving accelerators will likely involve increasingly heterogeneous architectures, combining general-purpose CPUs and GPUs, task-specific accelerators of different types like PIM-based cores and some programmable hardware like FPGAs or CGRAs for long-term adaptability. The balance between specialization and flexibility will be crucial in designing these multi-core SoCs that can efficiently handle both current and future workloads.

\section{Conclusion}
Driven by advances in deep learning algorithms, autonomous driving systems have made significant progress over the past decade. However, achieving full autonomy, with its substantial demands, requires careful co-design across the entire system stack. In this survey, we provided an overview of the critical components of autonomous driving systems, highlighting recent advancements from both academia and industry across software and hardware. Given the increasing demands of higher autonomy levels, exemplified by the diverse computational and memory requirements of three state-of-the-art end-to-end systems, we discussed potential directions for future hardware platforms. We emphasized that sustainable, adaptable, and efficient hardware solutions will be essential to support the next generation of fully autonomous vehicles.

\section{Acknowledgment}
This work was supported in part by SSF, Swedish Foundation for Strategic Research and in part by Lise Meitner’s Grants for Israeli-Swedish Research Collaboration under grant number SIP21-0087.

\bibliographystyle{ieeetr}
\bibliography{ref}

\vspace{12pt}

\end{document}